\documentclass[accepted]{uai2025} 
                        
\usepackage[american]{babel}
\usepackage{amssymb}
\usepackage{hyperref}
\usepackage{algorithm}
\usepackage{algorithmic}
\usepackage{amsmath}
\usepackage{algorithm}
\usepackage{algorithmic}
\usepackage{graphicx}
\usepackage{subcaption}
\usepackage{mathtools}
\usepackage{amsthm}
\usepackage{bm}
\usepackage{comment}
\usepackage{booktabs} 
\usepackage{multirow} 
\usepackage{natbib} 
\usepackage{mathtools} 
\usepackage{booktabs} 
\usepackage{tikz} 

\title{Variational Learning of Gaussian Process Latent Variable Models through Stochastic Gradient Annealed Importance Sampling}

\author[1]{Jian Xu}
\author[2]{Shian Du}
\author[1]{Junmei Yang}
\author[1]{Qianli Ma}
\author[1]{Delu Zeng\thanks{Corresponding author: \texttt{dlzeng@scut.edu.cn}}}
\author[3]{John Paisley}

\affil[1]{%
   South China University of Technology, Guangzhou, China
}
\affil[2]{%
    Tsinghua University, Shenzhen, China
}
\affil[3]{%
   Columbia University, NY, USA
}

\begin{document}
\maketitle

\begin{abstract}
Gaussian Process Latent Variable Models (GPLVMs) have become increasingly popular for unsupervised tasks such as dimensionality reduction and missing data recovery due to their flexibility and non-linear nature. An importance-weighted version of the Bayesian GPLVMs has been proposed to obtain a tighter variational bound. However, this version of the approach is primarily limited to analyzing simple data structures, as the generation of an effective proposal distribution can become quite challenging in high-dimensional spaces or with complex data sets.  In this work, we propose VAIS-GPLVM, a variational Annealed Importance Sampling method that leverages time-inhomogeneous unadjusted Langevin dynamics to construct the variational posterior. By transforming the posterior into a sequence of intermediate distributions using annealing, we combine the strengths of Sequential Monte Carlo samplers and VI to explore a wider range of posterior distributions and gradually approach the target distribution. We further propose an efficient algorithm by reparameterizing all variables in the evidence lower bound (ELBO). Experimental results on both toy and image datasets demonstrate that our method outperforms state-of-the-art methods in terms of tighter variational bounds, higher log-likelihoods, and more robust convergence.
\end{abstract}
\section{Introduction}
Gaussian processes (GPs) \cite{rasmussen2003gaussian} have become a popular method for function estimation due to their non-parametric nature, flexibility, and ability to incorporate prior knowledge of the function. Gaussian Process Latent Variable Models (GPLVMs), introduced by \cite{lawrence2005probabilistic}, have paved the way for GPs to be utilized for unsupervised learning tasks such as dimensionality reduction and structure discovery for high-dimensional data. It provides a probabilistic mapping from an unobserved latent space $\mathbf{H}$ to data-space $\mathbf{X}$.

The work by \cite{titsias2010bayesian} proposed a Bayesian version of GPLVMs and introduced a variational inference (VI) framework for training GPLVMs using sparse representations to reduce model complexity. This method utilizes an approximate surrogate estimator $g(\mathbf{X},\mathbf{H})$  to replace the true  probability term $p(\mathbf{X})$, i.e. $\mathbb{E}_{q(\mathbf{H})}\left[ g(\mathbf{X},\mathbf{H})\right]=p(\mathbf{X})$. VI typically defines an evidence lower bound (ELBO) as the loss function for the model in place of log $p(\mathbf{X})$. To describe the accuracy of this lower bound, we discuss a Taylor expansion of log $p(\mathbf{X})$,
\begin{equation}
    \mathbb{E}_{q(\mathbf{H})}\left[\text{log}\,g(\mathbf{X},\mathbf{H})\right] \approx \text{log}\,p(\mathbf{X})-\frac{1}{2}\text{var}_{q(\mathbf{H})}\left[\frac{g(\mathbf{X},\mathbf{H})}{p(\mathbf{X})}\right]
\end{equation}
The formula has been discussed in numerous works, including \cite{thin2020metflow,maddison2017filtering,domke2018importance}. Therefore, as the variance of the estimator decreases, the ELBO becomes tighter. Based on this formula and the basic principles of the central limit theorem, importance-weighted (IW) VI \cite{domke2018importance} seeks to reduce the variance of the estimator by repeatedly sampling from the proposal distribution $q(\mathbf{H})$, i.e., 
$
g\left( \mathbf{X},\mathbf{H} \right) =\frac{1}{K}\sum\nolimits_{k=1}^K{\left[ \frac{p\left( \mathbf{X}, \mathbf{H}_k \right)}{q\left( \mathbf{H}_k \right)} \right]}, \text{where}\, \mathbf{H}_k\sim q\left( \mathbf{H}_k \right) 
$. An importance-weighted version \cite{salimbeni2019deep} of the Bayesian GPLVMs based on this has  been proposed to obtain a tighter variational bound. While this method can obtain a tighter lower bound than the classical VI, it is a common problem  that the relative variance of this importance-sampling based estimator tends to increase with the dimension of the latent variable. Moreover, the generation of an effective proposal distribution can become quite challenging in high-dimensional spaces or with complex data sets. The problem of standard importance sampling techniques is that it can be challenging to construct a proposal distribution $q(\mathbf{H})$ that performs well in high-dimensional spaces, as shown in \cite{rainforth2018tighter, rudner2021signal}. 

To address these limitations, we propose VAIS-GPLVM, a variational Annealed Importance Sampling method that leverages time-inhomogeneous unadjusted Langevin dynamics to construct the variational posterior. Our method builds on the foundations of AIS, originally derived from nonequilibrium statistical mechanics \cite{jarzynski1997nonequilibrium}, and later extended in \cite{crooks1998nonequilibrium, neal2001annealed}. AIS remains a gold-standard technique for unbiased evidence estimation, as it explores a broader range of posterior distributions and gradually approaches the target distribution \cite{del2006sequential, salimans2015markov, grosse2013annealing, grosse2015sandwiching}.

Specifically, our proposed approach leverages an annealing procedure to transform the posterior distribution into a sequence of intermediate distributions, which can be approximated by using a Langevin stochastic flow. This dynamic is a time-inhomogeneous unadjusted Langevin dynamic that is easy to sample and optimize. We also propose an efficient algorithm designed by reparameterizing all variables in the  ELBO. Furthermore, we propose a stochastic variant of our algorithm that utilizes gradients estimated from a subset of the dataset, which improves the speed and scalability of the algorithm . Our experiments on both toy and image datasets show that our approach outperforms state-of-the-art methods in GPLVMs, demonstrating lower variational bounds, higher log-likelihoods, and more robust convergence.

Overall, our contributions are as follows:
\begin{itemize}
    \item  We propose VAIS-GPLVM, a variational Annealed Importance Sampling method that uses time-inhomogeneous unadjusted Langevin dynamics to construct the variational posterior. This approach mitigates the issue of weight collapse in high-dimensional GPLVMs, yielding a tighter lower bound and improved variational approximation for complex, high-dimensional data.

    \item We propose an efficient algorithm designed by reparameterizing all variables to further improve the estimation of the variational lower bounds. We also leverage stochastic optimization to maximize optimization efficiency.
    
    \item Our experiments on both toy and image datasets demonstrate that our approach outperforms state-of-the-art methods in GPLVMs, showing lower variational bounds, higher log-likelihoods, and more robust convergence.
\end{itemize}
\section{Background}

\subsection{GPLVM Variational Inference}
In GPLVMs, we have a training set comprising of $N$ $D$-dimensional real valued observations $\mathbf{X} = \{\boldsymbol{x}_n\}_{n=1}^N \in \mathbb{R}^{N \times D}$. These data are associated with $N$ $Q$-dimensional latent variables, $\mathbf{H} = \{\boldsymbol{h}_n\}_{n=1}^N \in \mathbb{R}^{N \times Q}$ where $Q < D$ provides dimensionality reduction \cite{titsias2010bayesian}. The forward mapping $\mathbf{H} \to \mathbf{X}$ is described by  multi-output GPs independently defined across dimensions $D$. The work by \cite{titsias2010bayesian} proposed a Bayesian version of GPLVMs  using sparse representations to reduce model complexity. We typically define the conditional distribution as $
p(\boldsymbol{f}_{d} \mid \boldsymbol{u}_{d}, \mathbf{H}) = \mathcal{N}\left(\boldsymbol{f}_{d}; \boldsymbol{\mu}_d, Q_{nn}\right)$, where $\boldsymbol{\mu}_d = K_{nm} K_{mm}^{-1} \boldsymbol{u}_d$, $Q_{nn} = K_{nn} - K_{nm} K_{mm}^{-1} K_{mn}$, $\boldsymbol{u}_d$ is the inducing variable \cite{titsias2009variational}.
Here, $K_{nn}$ is the covariance matrix evaluated over latent inputs $\{\boldsymbol{h}_n\}_{n=1}^N$ using a user-defined positive-definite kernel function $k_\theta(\boldsymbol{h}, \boldsymbol{h}')$, parameterized by a shared set of kernel hyperparameters $\theta$ across all output dimensions $D$.
The data likelihood is modeled as a Gaussian distribution, i.e.,

\begin{equation}
\begin{aligned}
p(\mathbf{X} \mid \mathbf{F}, \mathbf{H}) & =\prod_{n=1}^{N} \prod_{d=1}^{D} \mathcal{N}\left(x_{n, d} ; \boldsymbol{f}_{d}\left(\boldsymbol{h}_{n}\right), \sigma^{2}\right)
\end{aligned}
\end{equation}
where $\mathbf{F} = \{\boldsymbol{f}_d\}_{d=1}^{D}$, $\boldsymbol{x}_d$  is the $d$-th column of $\mathbf{X}$, and $m$ is  the number of inducing points. It is assumed that the prior is defined as $p(\boldsymbol{u}_d)=\mathcal{N}(\boldsymbol{0},K_{mm})$ and $p(\boldsymbol{h}_n)=\mathcal{N}(\boldsymbol{0},I_Q)$. Since $\boldsymbol{h}_n \in \mathbb{R}^Q$ is unobservable, we need to do joint inference over $\boldsymbol{f}(\cdot)$ and $\boldsymbol{h}$. Under the typical mean-field assumption of a factorized approximate posterior $q(\boldsymbol{f}_d)q(\boldsymbol{h}_n)$. We denote $\psi$ as all variational parameters and $\gamma$ as all GP hyperparameters. Thus, we arrive at the classical Mean-Field (MF) ELBO:
\begin{align}
    \begin{aligned}
&\operatorname{MF-ELBO}(\gamma, \psi)=\\&\sum_{n=1}^{N}\sum_{d=1}^{D} (\int q(\boldsymbol{f}_d) q\left(\boldsymbol{h}_{n}\right) \log p\left(x_{n,d} \mid  \boldsymbol{f}_d, \boldsymbol{h}_{n}\right) \mathrm{d} \boldsymbol{h}_{n} \mathrm{~d} \boldsymbol{f}_d \\
&-\operatorname{KL}\left(q\left(\boldsymbol{h}_{n}\right) \| p\left(\boldsymbol{h}_{n}\right)\right)-\operatorname{KL}\left(q(\boldsymbol{u}_d) \| p(\boldsymbol{u}_d)\right)),
\end{aligned}
\end{align}
where we use the typical approximation to integrate out the inducing variable,
\begin{equation}
\label{conditional}
   q\left( \boldsymbol{f}_d \right) =\int{p\left( \boldsymbol{f}_d|\boldsymbol{u}_d \right)}q\left( \boldsymbol{u}_d \right) d\boldsymbol{u}_d.
\end{equation}

In Equation (\ref{conditional}), $p\left( \boldsymbol{f}_d|\boldsymbol{u}_d \right)$
 is a simplification of the traditional Sparse Gaussian Process (Sparse GP) approach. In the Sparse GP model, we typically assume $p\left( \boldsymbol{f}_d|\boldsymbol{u}_d, \boldsymbol{h}_n \right)$
 as the conditional probability distribution of the latent variable $\boldsymbol{h}_n$
, and we integrate over $\boldsymbol{h}_n$. Proofs can be seen in the Appendix.

\subsection{Importance-weighted Variational Inference}
\label{IW}
A main contribution of \cite{salimbeni2019deep} is to propose a variational scheme for LV-GP models based on importance-weighted VI \cite{domke2018importance} via amortizing the optimization of the local variational parameters. IWVI provides a way of lower-bounding the log marginal likelihood more tightly
and with less estimation variance by Jensen’s inequality at the expense of increased computational complexity. The IW-ELBO is obtained by replacing the expectation likelihood term in Vanilla VI with a sample average of $K$ terms: 
\begin{align}
\label{(8)}
    \operatorname{IW-ELBO}(\gamma, \psi)= \sum_{n=1}^{N} \sum_{d=1}^{D}(B_{n,d}-\operatorname{KL}\left(q(\boldsymbol{u}_d) \| p(\boldsymbol{u}_d)\right)),
\end{align}
where $
B_{n,d}=\underset{\boldsymbol{f}_d,\boldsymbol{h}_n}{\mathbb{E}}\log \frac{1}{K}\sum_k{p}\left( x_{n,d}\mid \boldsymbol{f}_d,\boldsymbol{h}_{n,k} \right) \frac{p\left( \boldsymbol{h}_{n,k} \right)}{q\left( \boldsymbol{h}_{n,k} \right)}
$.  Proofs can be seen in the Appendix.

Although the IW objective outperforms classical VI in terms of accuracy, its effectiveness is contingent on the variability of the importance weights: $p\left( x_{n,d}\mid \boldsymbol{f}_d,\boldsymbol{h}_{n,k} \right) \frac{p\left( \boldsymbol{h}_{n,k} \right)}{q\left( \boldsymbol{h}_{n,k} \right)}$. When these
weights vary widely, the estimate will effectively rely on only the few points with
the largest weights, as shown in \cite{rainforth2018tighter}. To ensure the effectiveness of importance sampling, the proposal distribution defined by $
q\left( \boldsymbol{h}_{n,k} \right) 
$ must therefore be a fairly good approximation to $
p\left( x_{n,d}\mid \boldsymbol{f}_d,\boldsymbol{h}_{n,k} \right) p\left( \boldsymbol{h}_{n,k} \right) 
$, so that the importance weights do not vary widely. Related theoretical proofs can be seen in 
 \cite{domke2018importance,maddison2017filtering,rainforth2018tighter}.

When $\boldsymbol{h}_{n,k}$ is high-dimensional, or the likelihood $
p\left( x_{n,d}\mid \boldsymbol{f}_d,\boldsymbol{h}_{n,k} \right) 
$ is  multi-modal, finding a good importance sampling distribution can be very difficult, limiting the applicability of the method. Unfortunately, original research by \cite{salimbeni2019deep} only discusses the case when $\boldsymbol{h}_n$ is a one-dimensional latent variable, and they acknowledge that reliable inference for more complex cases is not yet fully understood or documented. To circumvent this issue, we provide an alternative for variational GPLVMs using Annealed Importance Sampling (AIS) \cite{crooks1998nonequilibrium,neal2001annealed,wu2016quantitative}, which defines state-of-the-art estimators of the evidence and designs efficient proposal importance distributions. Specially,
we propose a novel ELBO, relying on unadjusted Langevin dynamics, which is a simple implementation that  combines the strengths of Sequential Monte
Carlo  samplers  and variational inference as detailed in  Section \ref{method}.

\section{ Variational AIS Scheme in GPLVMs }
\label{method}
\subsection{Variational Inference via AIS}
\label{4.1}

Annealed Importance Sampling (AIS)\cite{neal2001annealed,del2006sequential,salimans2015markov} is a technique for obtaining an unbiased estimate of the evidence $p(\mathbf{X})$. To achieve this, AIS uses a sequence of $K$ bridging densities $\left\{q_{k}(\mathbf{H})\right\}_{k=1}^{K}$ that connect a simple base distribution $q_0(\mathbf{H})$ to the posterior distribution $p(\mathbf{H}|\mathbf{X})$. By gradually interpolating between these distributions, AIS allows for an efficient computation of the evidence. This method is particularly useful when the posterior is difficult to sample from directly, as it allows us to estimate the evidence without evaluating the full posterior distribution directly. We can express this as follows:
\begin{align}
    p(\mathbf{X})=\int  p(\mathbf{X}, \mathbf{H})d \mathbf{H}=\mathbb{E}_{q_{\mathrm{fwd}}\left(\mathbf{H}_{0: K}\right)}\left[\frac{q_{\mathrm{bwd}}\left(\mathbf{H}_{0: K}\right)}{q_{\mathrm{fwd}}\left(\mathbf{H}_{0: K}\right)}\right]
\end{align}
where the variational distribution $q_{\mathrm{fwd}}$ and the 
target distribution $q_{\mathrm{bwd}}$ can be written as:
\begin{small}
\begin{align}
\label{10}
    \begin{aligned}
q_{\mathrm{fwd}}\left(\mathbf{H}_{0: K}\right) &=q_{0}\left(\mathbf{H}_{0}\right) \mathcal{T}_{1}\left(\mathbf{H}_{1} \mid \mathbf{H}_{0}\right) \cdots \mathcal{T}_{K}\left(\mathbf{H}_{K} \mid \mathbf{H}_{K-1}\right) \\
q_{\mathrm{bwd}}\left(\mathbf{H}_{0: K}\right) &=p\left(\mathbf{X}, \mathbf{H}_{K}\right) \tilde{\mathcal{T}}_{K}\left(\mathbf{H}_{K-1} \mid \mathbf{H}_{K}\right) \cdots \tilde{\mathcal{T}}_{1}\left(\mathbf{H}_{0} \mid \mathbf{H}_{1}\right)
\end{aligned}
\end{align}
\end{small}
Here, we assume $\mathcal{T}_{k}$ is a forward MCMC kernel that leaves $q_{k}(\mathbf{H})$ invariant, which ensures that $\{ \mathcal{T}_{k}\}_{k=1}^{K}$ are valid transition probabilities, i.e., \begin{align}\int q_{k}(\mathbf{H}_{k-1}) \mathcal{T}_{k}\left(\mathbf{H}_k\mid  \mathbf{H}_{k-1}\right) \mathrm{d} \mathbf{H}_{k-1}=q_{k}\left(\mathbf{H}_{k}\right).\end{align} And $\tilde{\mathcal{T}}_{k}$ is the “backward” Markov kernel moving each sample $\mathbf{H}_k$ into a sample $\mathbf{H}_{k-1}$ starting from a virtual sample
$\mathbf{H}_K$. $q_{\mathrm{fwd}}$ represents the chain of states generated by AIS, and $q_{\mathrm{bwd}}$ is a fictitious
reverse chain which begins with a sample from $p(\mathbf{X}, \mathbf{H})$ and applies the transitions in reverse order. In practice, the bridging densities have to be chosen carefully for a low variance
estimate of the evidence. A typically method is to use geometric averages of the initial and target distributions to construct the sequence, i.e., $q_{k}(\mathbf{H}) \propto q_{0}(\mathbf{H})^{1-\beta_{k}} p(\mathbf{X}, \mathbf{H})^{\beta_{k}}$ for $ 0=\beta_{0}<\beta_{1}<\cdots<\beta_{K}=1$. AIS has been proven theoretically to be consistent as $K \rightarrow \infty$ \cite{neal2001annealed} and
achieves accurate estimate of $\log p(\mathbf{X})$
empirically with the asymptotic bias decreasing at a $1/K$ rate \cite{grosse2013annealing,grosse2015sandwiching}.

\begin{figure}[ht]
\centering
\includegraphics[width=0.9\linewidth]{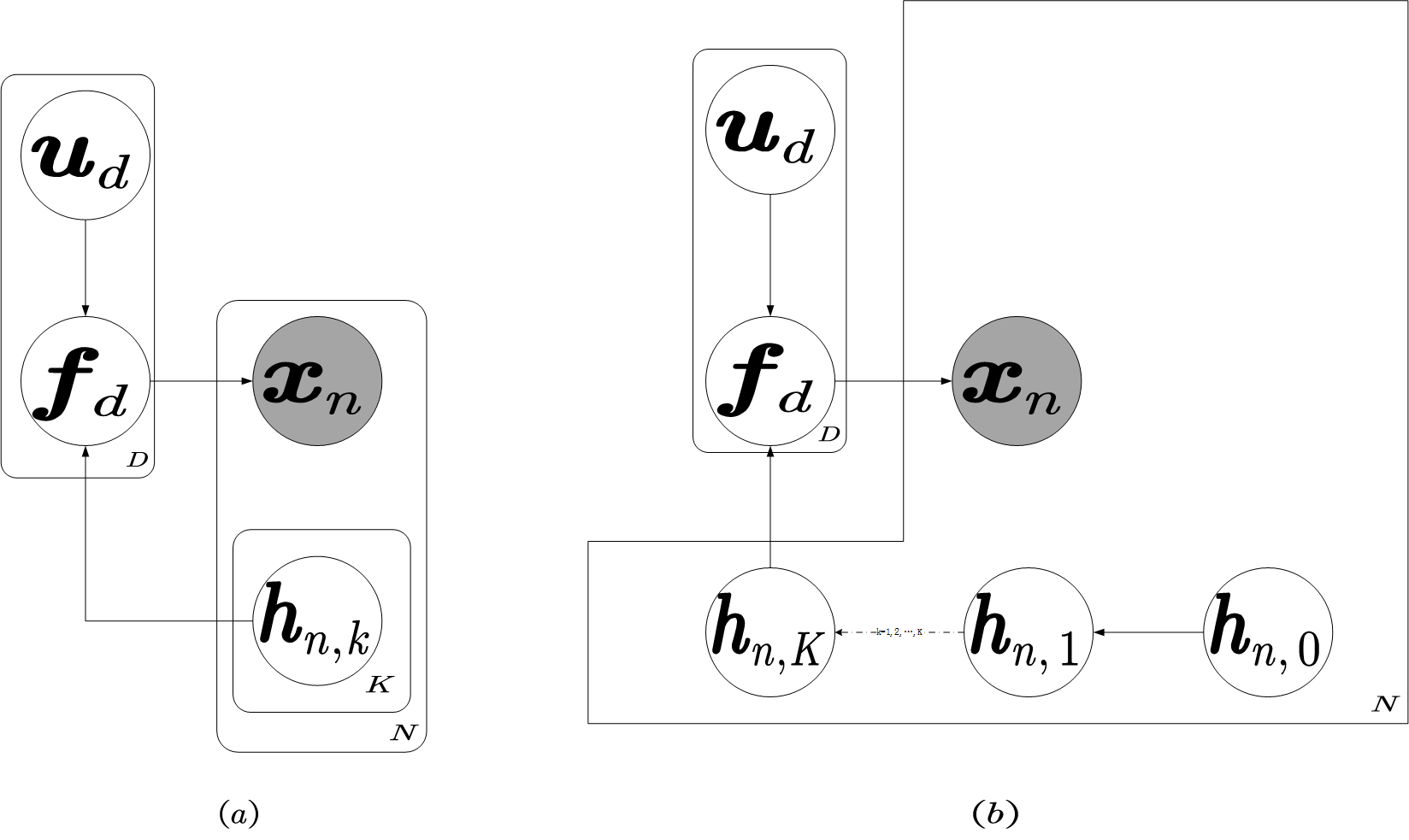}
\caption{ The graphical models of (a) IW and (b) our method. We leverages an annealing procedure to transform the posterior distribution into a
sequence of intermediate distributions.}
\label{fig:3}
\end{figure}

With this, we can derive the AIS bound,
\begin{align}
\label{12}
\begin{aligned}
\log p(\mathbf{X})&\ge \mathbb{E}_{q_{\mathrm{fwd}}\left(\mathbf{H}_{0: K}\right)}\left[\log\frac{q_{\mathrm{bwd}}\left(\mathbf{H}_{0: K}\right)}{q_{\mathrm{fwd}}\left(\mathbf{H}_{0: K}\right)}\right]\\&=\mathbb{E}_{q_{\mathrm{fwd}\left(\mathbf{H}_{0: K}\right)}}[\log p\left(\mathbf{X}, \mathbf{H}_{K} \right)-\log q_{0}\left(\mathbf{H}_{0}\right)\\&-\sum_{k=1}^{K} \log \frac{\mathcal{T}_{k}(\mathbf{H}_{k}\mid\mathbf{H}_{k-1})}{\tilde{\mathcal{T}}_{k}\left(\mathbf{H}_{k-1}\mid\mathbf{H}_{k} \right)}].
\end{aligned}
\end{align}
This objective can be obtained by applying Jensen’s inequality. For the GPLVM model, we  can naturally derive its AIS lower bound:
\begin{align}
\label{13}
\begin{aligned}
&\mathcal{L} _{\mathrm{AIS}}(\psi ,\gamma )\\=&\sum_{n=1}^N{\sum_{d=1}^D{\mathbb{E} _{q_{\mathrm{fwd}}\left( \boldsymbol{h}_{0:K} \right) q\left( \boldsymbol{f}_{\boldsymbol{d}} \right)}\left[ \log p\left( x_{n,d}\mid \boldsymbol{f}_d,\boldsymbol{h}_{n,K} \right) \right]}}
\\
&+\sum_{n=1}^N{\mathbb{E} _{q_{\mathrm{fwd}}\left( \boldsymbol{h}_{0:K} \right)}\left[ \log p\left( \boldsymbol{h}_{n,K} \right) -\log q_0\left( \boldsymbol{h}_{n,0} \right) \right]}
\\
&-\sum_{k=1}^K{\mathbb{E} _{q_{\mathrm{fwd}}\left( \mathbf{H}_{0:K} \right)}\log \frac{\mathcal{T} _k\left( \mathbf{H}_k\mid \mathbf{H}_{k-1} \right)}{\tilde{\mathcal{T}}_k\left( \mathbf{H}_{k-1}\mid \mathbf{H}_k \right)}}\\&-\sum_{d=1}^D{\mathrm{KL}\left( q(\boldsymbol{u}_d)\parallel p(\boldsymbol{u}_d) \right)}
\end{aligned}
\end{align}
where $\psi$ and $\gamma$ indicate the sets of all variational parameters and all GP hyperparameters, respectively. Our purpose is to evaluate this bound. First we note that the
last $\mathrm{KL}$ term is tractable if we assume the variational posteriors of $\boldsymbol{u_d}$  are mean-field Gaussian distributions. So we concentrate on the terms in  the expectation that we can evaluate relying on a Monte Carlo estimate. It is obvious that $
\log p\left( x_{n,d}\mid \boldsymbol{f}_d,\boldsymbol{h}_{n,K} \right) 
$ is available in closed form as the conditional likelihood is Gaussian \cite{titsias2009variational}. Therefore, the first  three term can be computed 
by the popular ``reparameterization trick'' \cite{rezende2014stochastic,kingma2013auto} to obtain an unbiased estimate of the expectation over $
q_{\mathrm{fwd}}\left( \mathbf{H}_{0:K} \right)$ and $ q\left( \boldsymbol{f}_{\boldsymbol{d}} \right) 
$ (detailed in Section \ref{3.3}). 
Afterwards, to evaluate expectation over $q_{\mathrm{fwd}}$, we construct an MCMC transition operator $\mathcal{T}_k$ which leaves $q_k$ invariant via a  time-inhomogeneous unadjusted
(overdamped) Langevin algorithm (ULA) as used in \cite{welling2011bayesian,heng2020controlled,wu2020stochastic,marceau2017natural} and jointly optimize $\psi$ and $\gamma$ by stochastic gradient descent. For visualization, we present our AIS method alongside the traditional IW method's graphical model in Fig. \ref{fig:3}.

\begin{algorithm*}[ht]
\caption{Stochastic  Unadjusted Langevin Diffusion (ULA)
 AIS algorithm for GPLVMs }
\label{algorithm}
\begin{algorithmic}
   \STATE {\bfseries Input:} training data $\mathbf{X}$,
   mini-batch size $B$, sample number $K$, annealing schedule $\left\{\beta_{k}\right\}$, stepsizes $\eta$
\STATE {\bfseries Initialize}  all GPLVM hyperparameters $\gamma$, all variational parameters $\psi$
   \REPEAT
   \STATE Sample  mini-batch indices $J \subset\{1, \ldots, N\} \text { with }|J|=B$ 
   \STATE Draw $
\boldsymbol{\epsilon}$ from  standard Gaussian distribution.
   \STATE Set $\mathbf{H}_{0}=\mathbf{a_n}+ {L_n}\boldsymbol{\epsilon}$
   \STATE Set $\mathcal{L}=-\log q_0(\mathbf{H}_{0})$
   \FOR{$k=1$ {\bfseries to} $K$}
   
   \STATE Draw $
\boldsymbol{\epsilon} _k$ from  standard Gaussian distribution.
   
   \STATE Set $\nabla \log q_{k}\left(\cdot\right)=\beta_{k}\nabla (\frac{N}{B}\log p\left( \mathbf{X}_J|\cdot \right) +\log p\left( \cdot \right))+\left(1-\beta_{k}\right) \nabla\log q_{0}(\cdot )$
   \STATE Set $\mathbf{H}_{k}=\mathbf{H}_{k-1}+\eta \nabla \log q_{k}\left(\mathbf{H}_{k-1}\right)+\sqrt{2 \eta} \boldsymbol{\epsilon} _{k-1}$
   \STATE Set $\tilde{\boldsymbol{\epsilon}}_{k-1}=\sqrt{\frac{\eta}{2}}\left[\nabla \log q_{k}\left(\mathbf{H}_{k-1}\right)+\nabla \log q_{k}\left(\mathbf{H}_{k}\right)\right]-\boldsymbol{\epsilon}_{k-1}$
   \STATE Set $R_{k-1}=\frac{1}{2}\left(\left\|\tilde{\boldsymbol{\epsilon}}_{k-1}\right\|^{2}-\left\|\boldsymbol{\epsilon}_{k-1}\right\|^{2}\right)$
   \STATE Set $\mathcal{L}=\mathcal{L}-R_{k-1}$

   \ENDFOR
   \STATE Sample  mini-batch indices $I \subset\{1, \ldots, N\} \text { with }|I|=B$ 
   \STATE  Draw $
\epsilon _{\boldsymbol{f}_d}$ from  standard Gaussian distribution for $d=1,2,...,D$ .
    \STATE Set $
\mathcal{L}=\mathcal{L}+ \log p\left(\mathbf{H}_{K}\right)+\frac{N}{B}\log p\left( \mathbf{X}_I\mid \boldsymbol{\epsilon }_{f_d},\boldsymbol{\epsilon }_{0:K-1},\boldsymbol{\epsilon } \right)-\sum_{d=1}^D{\mathrm{KL}\left( q(\boldsymbol{u}_d)\parallel p(\boldsymbol{u}_d) \right)}$ 
    \STATE Do gradient descent on $\mathcal{L}(\psi ,\gamma)$

    \UNTIL{$\psi,\gamma$  converge}
    \end{algorithmic}
\end{algorithm*}
\subsection{Time-inhomogeneous Unadjusted Langevin Diffusion}
$\mathcal{T}_{k}$ can be constructed using a Markov kernel with an invariant density such as MH or HMC, which enables $q_{\mathrm{fwd}}$ to converge to the posterior distribution of $\mathbf{H}$. For the sake of simplicity, we consider  the transition density $\mathcal{T}_{k}$ associated to this discretization,
\begin{align}
\begin{aligned}
\label{14}
&\mathcal{T}_{k}\left(\mathbf{H}_{k}\mid \mathbf{H}_{k-1}\right)\\=&~\mathcal{N}\left(\mathbf{H}_{k} ; \mathbf{H}_{k-1}+\eta \nabla \log q_{k}\left(\mathbf{H}_{k-1}\right), 2 \eta I\right) 
\end{aligned}
\end{align}
where $\eta> 0 $ is the step size and $q_{k}$ is bridging densities defined in Section \ref{4.1}. Since  we have $q_{k}(\mathbf{H}) \propto q_{0}(\mathbf{H})^{1-\beta_{k}} p(\mathbf{X}, \mathbf{H})^{\beta_{k}}$ in Section \ref{4.1}, the annealed potential energy is derived as:
\begin{align}
\label{15}
    \nabla \log q_{k}\left(\cdot\right)=\beta_{k} \nabla\log p(\mathbf{X},\cdot)+\left(1-\beta_{k}\right) \nabla\log q_{0}(\cdot ).
\end{align}
According to conditional probability formula $
\log p(\mathbf{X},\cdot )=\log p\left( \mathbf{X}|\cdot \right) +\log p\left( \cdot \right) $, the model log likelihood simplifies to: 
\begin{align}
\label{16}
\begin{aligned}
 &\nabla \log p(\mathbf{X}|\cdot )=
-\frac{1}{2}\sum_{d=1}^D\nabla ( \log  \det \left( Q_{nn}+\sigma ^2I \right)\\& +\left( \boldsymbol{x}_d-\boldsymbol{\mu_d} \right) ^T\left( Q_{nn}+\sigma ^2I \right) ^{-1}\left( \boldsymbol{x}_d-\boldsymbol{\mu_d} \right) ).
\end{aligned}
\end{align}
Since Eq. (\ref{16}) is analytical, the gradient can be computed through automatic differentiation \cite{baydin2018automatic}.
The dynamical system propagates from  a base variational distribution $q_0$ to a final distribution $q_K$ which approximates the posterior density. Let $\eta := T /K$, then
the proposal  $q_{\mathrm{fwd}}$ converges to the path measure of the following  Langevin diffusion
$\left(\mathbf{h}_{t}\right)_{t \in[0, T]}$ defined by the stochastic differential equation (SDE),
\begin{align}
    \mathbf{dH}_{t}=\nabla \log q_{t}( \mathbf{H}) \mathrm{d} t+\sqrt{2} \mathrm{~d} \mathbf{B}_{t}, \quad \mathbf{H}_{0} \sim q_{0}
\end{align}
where $\left(\mathbf{B}_{t}\right)_{t \in[0, T]}$
is standard multivariate Brownian motion and $q_t$ corresponds to $q_k$ in discrete-time for $t=t_{k}=k \eta$. For long times, the
solution of the  Fokker-Planck equations \cite{risken1996fokker} tends to the stationary distribution  $q_\infty (\mathbf{H}) \propto \exp (p(\mathbf{X}, \mathbf{H}))$. Additional quantitative results measuring the law of $\mathbf{h}_T$  for such annealed diffusions have been showed in \cite{andrieu2016sampling,tang2021simulated,fournier2021simulated}.
For ease of sampling, we define the corresponding Euler-Maruyama discretization as,
\begin{align}
\label{ula}
    \mathbf{H}_{k}=\mathbf{H}_{k-1}+\eta \nabla \log q_{k}\left(\mathbf{H}_{k-1}\right)+\sqrt{2 \eta} \boldsymbol{\epsilon} _{k-1},
\end{align}
where $\boldsymbol{\epsilon}_k \sim \mathcal{N}(0,I)$, as done in \cite{heng2020controlled,wu2020stochastic,nilmeier2011nonequilibrium}. Since such process is reversible w.r.t. $q_k$, based on \cite{nilmeier2011nonequilibrium}, the reversal $\tilde{\mathcal{T}}_{k}$ is typically realized by,
\begin{align}
\mathbf{H}_{k-1}=\mathbf{H}_k+\eta \nabla \log q_k\left( \mathbf{H}_k \right) +\sqrt{2\eta}\tilde{\boldsymbol{\epsilon}}_{k-1},
\end{align}
where $\tilde{\boldsymbol{\epsilon}}_{k-1}=-\sqrt{\frac{\eta}{2}}\left[\nabla \log q_{k}\left(\mathbf{H}_{k-1}\right)+\nabla \log q_{k}\left(\mathbf{H}_{k}\right)\right]-\boldsymbol{\epsilon}_{k-1}$. Based on Eq. (\ref{14}), the term related to $\mathcal{T}_{k}$ in Eq. (\ref{13}) can be written explicitly as:
\begin{align}
\label{20}
\begin{aligned}
\sum_{k=1}^{K}R_{k-1}&=\sum_{k=1}^{K} \log 
\frac{\mathcal{T} _k\left( \mathbf{H}_k\mid \mathbf{H}_{k-1} \right)}{\tilde{\mathcal{T}}_k\left( \mathbf{H}_{k-1}\mid \mathbf{H}_k \right)}\\&=\sum_{k=1}^{K}\frac{1}{2}\left(\left\|\tilde{\boldsymbol{\epsilon}}_{k-1}\right\|^{2}-\left\|\boldsymbol{\epsilon}_{k-1}\right\|^{2}\right).     
\end{aligned}
\end{align}
We abbreviate this probability ratio as $R_{k-1}$. Additional proofs can be seen in Appendix A. 
\begin{table*}[ht]
  \centering
  \caption{Comparison of MF, IW, and AIS under different number of iterations for two toy datasets}
  \label{tab:comparison}
  \resizebox{\textwidth}{!}{
  \begin{tabular}{ccccccc}
    \toprule
    Dataset & Data Dim & Method & Iterations & Negative ELBO & MSE & Negative Expected Log Likelihood \\
    \midrule
    \multirow{9}{*}{Oilflow} & \multirow{9}{*}{(1000,12)} & \multirow{3}{*}{MF-GPLVM} & 1000 & 3.44 (0.25) & 6.83 (0.27) & -1.42 (0.27) \\
       &    &  & 2000 & -1.67 (0.17) & 3.59 (0.13) & -8.38 (0.12) \\
       &    &  & 3000 & -3.07 (0.12) & 2.79 (0.11) & -11.24 (0.10) \\
       \cmidrule(lr){3-7}
       &  & \multirow{3}{*}{IWVI-GPLVM} & 1000 & \textbf{0.01} (0.25) & \textbf{4.52} (0.28) & \textbf{-6.26} (0.26) \\
       &    & & 2000 & -3.19 (0.15) & 2.77 (0.16) & -9.46 (0.15) \\
       &    &  & 3000 & -4.13 (0.14) & 2.60 (0.15) & -12.20 (0.12) \\
       \cmidrule(lr){3-7}
       &  & \multirow{3}{*}{VAIS-GPLVM (ours)} & 1000 & 0.78 (0.24) & 4.99 (0.23) & -4.01 (0.26) \\
       &    &  & 2000 & \textbf{-5.04} (0.15) & \textbf{2.65} (0.15) & \textbf{-10.33} (0.16) \\
       &    &  & 3000 & \textbf{-6.82} (0.12) & \textbf{2.16} (0.12) & \textbf{-13.06} (0.11) \\
       \midrule
    \multirow{9}{*}{Wine Quality} & \multirow{9}{*}{(1599,11)} & \multirow{3}{*}{MF-GPLVM} & 1000 & 32.69(0.13) & 63.98(0.12) & 31.71(0.15) \\
       &    &  & 2000 & 13.46(0.03) & 48.95(0.05) & 6.51(0.06) \\
       &    &  & 3000 & 11.59(0.03) & 45.81(0.04) & 4.07(0.05) \\
       \cmidrule(lr){3-7}
       & & \multirow{3}{*}{IWVI-GPLVM} & 1000 & \textbf{22.65}(0.07) & \textbf{50.77}(0.06) & \textbf{19.94}(0.09) \\
       &    &  & 2000 & 11.47(0.02) & 40.86(0.03) & 3.72(0.04) \\
       &    &  & 3000 & 10.73(0.03) &  35.23(0.04)& 2.71(0.03) \\
       \cmidrule(lr){3-7}
       &  & \multirow{3}{*}{VAIS-GPLVM (ours)} & 1000 & 29.63(0.07) &  57.49(0.05)& 27.67(0.06) \\
       &    &  & 2000 & \textbf{10.43}(0.03) & \textbf{34.60}(0.03) & \textbf{3.58}(0.04) \\
       &    & & 3000 & \textbf{8.86}(0.04) & \textbf{32.23}(0.04) & \textbf{2.47}(0.03) \\
    \bottomrule
  \end{tabular}}
\end{table*}
\subsection{Reparameterization Trick and Stochastic Gradient Descent}
\label{3.3}

For ease of sampling, we consider a reparameterization version of Eq. (\ref{13}) based on the Langevin mappings associated with $q_k$ given by
\begin{align}
T_{k }(\mathbf{H}_{k-1})=\mathbf{H}_{k-1}+\eta \nabla \log q_{k}\left(\mathbf{H}_{k-1}\right)+\sqrt{2 \eta} \boldsymbol{\epsilon}_{k-1}.  
\end{align}
Based on the identity $\mathbf{H}_{k} =T_{k}(\mathbf{H}_{k-1})$, we have a representation of $\mathbf{H}_{k}$ by a stochastic flow,
\begin{align}
 \mathbf{H}_{k}=T_{k}\left(\mathbf{H}_{k-1}\right)=T_{k}\circ T_{k-1}\circ\cdots T_{1}(\mathbf{H_0} )     
\end{align}
Moreover, for LVGP models, we also have a reparameterization version \cite{salimbeni2017doubly} of the posteriors of   $\mathbf{H_0}$ and $\boldsymbol{f}_d$ in Eq. (\ref{13}), that is, 
\begin{align}
\begin{aligned}
&\boldsymbol{h}_{n,0}=\boldsymbol{a}_{n}+L_{n}\boldsymbol{\epsilon}\\&
\boldsymbol{f}_d=K_{nm}K_{mm}^{-1}\boldsymbol{m}_d\\&+\sqrt{K_{nn}-K_{nm}K_{mm}^{-1}\left( K_{mm}-{\boldsymbol{S}_d}^T\boldsymbol{S}_d \right) K_{mm}^{-1}K_{mn}}\boldsymbol{\epsilon }_{f_d}
\end{aligned}
\end{align}
where vectors $\boldsymbol{a}_{n} \in \mathbb{R}^Q$, $\boldsymbol{m}_{d}\in \mathbb{R}^N$ and  upper triangular matrixs $L_{n}$, $\boldsymbol{S}_{d}$ are the variational parameters, $\boldsymbol{\epsilon} \in \mathbb{R}^Q,\boldsymbol{\epsilon}_{f_d} \in \mathbb{R}^N$ are standard Gaussian distribution. After this reparameterization, a change of variable  shows that AIS bound in Eq. (\ref{13}) can be rewritten as:
\begin{align}
\begin{aligned}
    &\mathcal{L} _{\mathrm{AIS}}(\psi ,\gamma )\\=&\sum_{n=1}^N{\sum_{d=1}^D{\begin{array}{c}
	\mathbb{E} _{p\left( \boldsymbol{\epsilon }_{f_d} \right) p\left( \boldsymbol{\epsilon }_{0:K-1} \right) p\left( \epsilon \right)}\left[ \log p\left( x_{n,d}\mid \boldsymbol{\epsilon }_{f_d},\boldsymbol{\epsilon }_{0:K-1},\boldsymbol{\epsilon } \right) \right] \,\,\\
\end{array}}}
\\
&+\sum_{n=1}^N{\mathbb{E} _{p\left( \boldsymbol{\epsilon }_{0:K-1} \right) p\left( \epsilon \right)}\left[ \log p\left( \boldsymbol{h}_{n,K} \right) -\log q_0\left( \boldsymbol{h}_{n,0} \right) \right]}
\\
&-\sum_{k=1}^K{\mathbb{E} _{p\left( \boldsymbol{\epsilon }_{0:K-1} \right) p\left( \boldsymbol{\epsilon } \right)}}R_{k-1}-\sum_{d=1}^D{\mathrm{KL}\left( q(\boldsymbol{u}_d)\parallel p(\boldsymbol{u}_d) \right)},
\\
\end{aligned}
\end{align}
where $
R_{k-1}
$ is defined in Eq. (\ref{20}) and $\mathbf{h}_{n,k}$ is reparameterized as $\mathbf{h}_{n,k}=T_{k} \circ T_{k-1} \circ \cdots T_{1}\left(\mathbf{h}_{n,0}\right)
=\bigcirc _{i=1}^{k} T_{i}(\mathbf{a}_{n}+ L_{n}\boldsymbol{\epsilon})$.

 In order to accelerate training and sampling in our inference scheme, we propose a  scalable variational bounds that are tractable in the large data regime based on stochastic variational inference \cite{hoffman2013stochastic,salimbeni2017doubly,kingma2013auto,hoffman2015structured,naesseth2020markovian} and stochastic gradient descent \cite{welling2011bayesian,chen2014stochastic,zou2019stochastic,teh2016consistency,sato2014approximation,alexos2022structured} as described in Algorithm
\ref{algorithm}.

Instead of computing the gradient of the full log likelihood, we suggest to use a stochastic variant to subsampling datasets into a mini-batch $\mathcal{D}_J$ with $\left | \mathbf{X}_J\ \right | =B$, where $J\subset \{1,2,..,N\}$ is the indice of
any mini-batch. In the meantime, we replace the $p\left(\mathbf{X}, \mathbf{H}_{K}\right)$ term in Eq. (\ref{10}) with another estimator computed using an independent mini-batch of indices  $I\subset \{1,2,..,N\}$
with $\left | \mathbf{X}_I\ \right | =B$. We finally derive a  stochastic variant of the Stochastic Unadjusted Langevin Diffusion AIS algorithm for the GPLVMs  as described in Algorithm \ref{algorithm}.

\begin{figure*}[t]
\centering
\includegraphics[width=\linewidth]{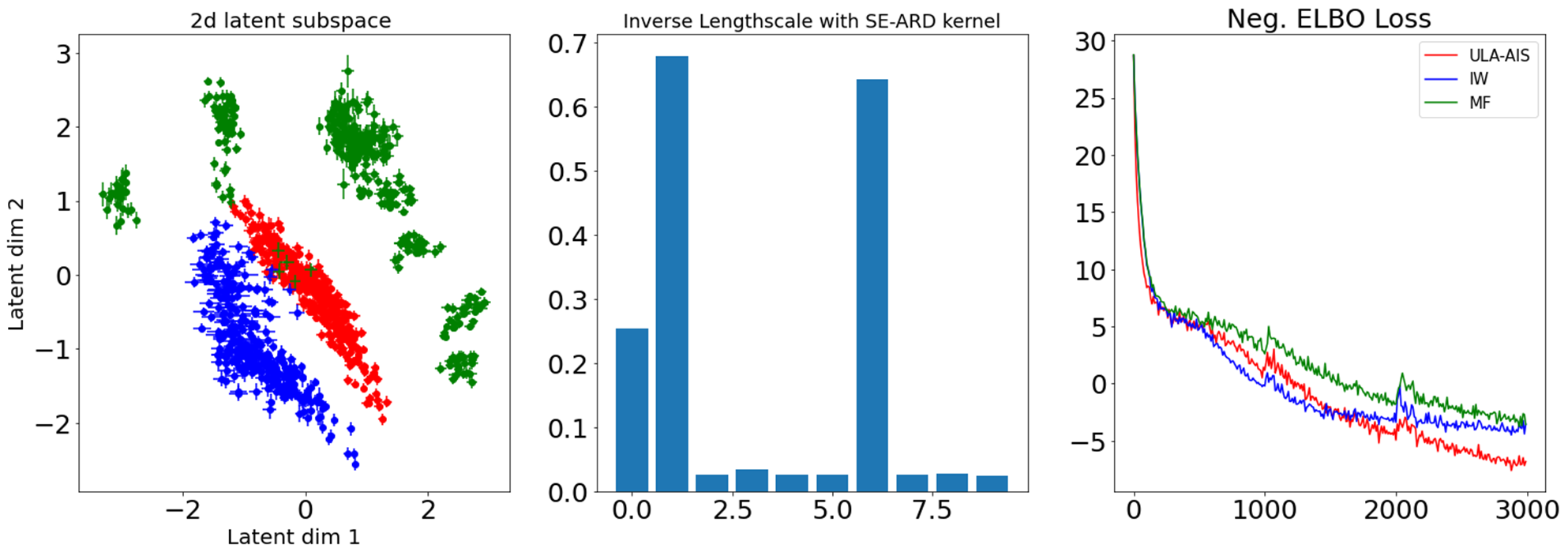}
\caption{We lowered the data dimensionality using our proposed method in the multi-phase oilflow dataset and visualized a two-dimensional slice of the latent space that corresponds to the most dominant latent dimensions. The inverse lengthscales learnt with SE-ARD kernel for each dimension  are depicted in the middle plot, and the negative ELBO learning curves are shown in the right plot. We set the same learning rate  and compared the learning curves of two state-of-the-art models, MF and Importance Weighted VI  within 3000 iterations for GPLVMs.}
\label{fig1}
\end{figure*}

\begin{figure}[ht]
\centering
\includegraphics[width=0.48\textwidth]{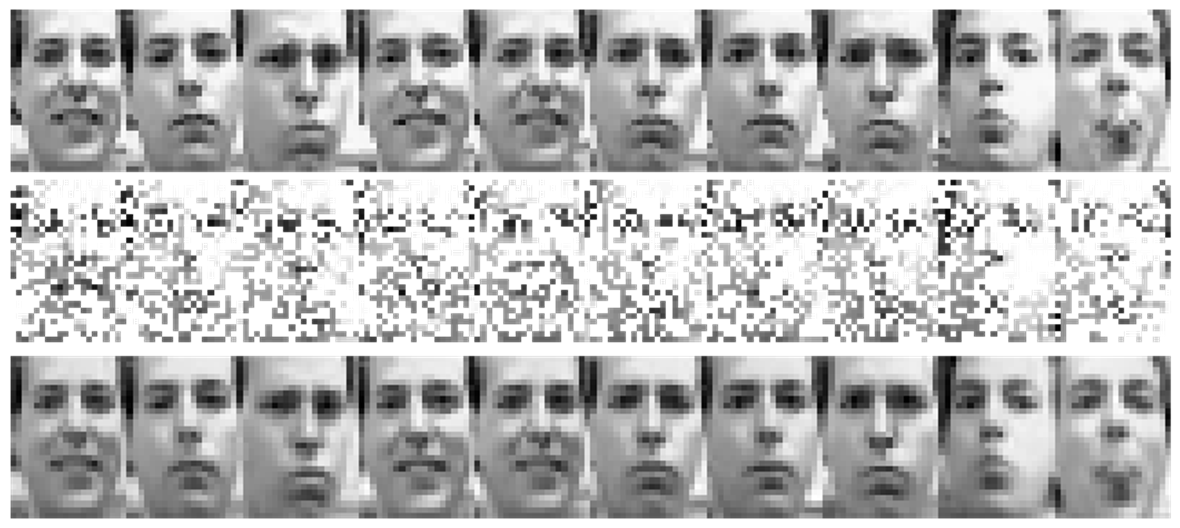}
\caption{In the Brendan faces reconstruction task with 75\% missing pixels, the top row represents the ground truth data and the bottom row showcases the reconstructions from the 20-dimensional latent distribution.
}
\label{fig:figure5}
\end{figure}

\section{Related Work}
\paragraph{IWVI}
IWVI~\cite{domke2018importance} demonstrated that importance weighting constitutes a form of augmented variational inference, thereby revealing the looseness inherent in previous variational objectives. This insight was later extended to the case of $\alpha$-divergences~\cite{geffnerempirical, daudel2023alpha}. However, \cite{rainforth2018tighter} showed that tighter ELBOs can reduce the gradient estimator’s signal-to-noise ratio (SNR), impairing inference network learning. Similarly, \cite{salimbeni2019deep} addressed this in Deep Gaussian Processes (DGPs) by introducing an importance-weighted objective with latent noisy covariates, balancing accuracy and computational cost through analytic solutions.

Building on this, \cite{rudner2021signal} found that increasing importance samples degrades gradient SNR for latent variable parameters, sometimes reducing gradients to pure noise. They mitigated this by adapting doubly-reparameterized gradient estimators to the DGP context, improving stability. In contrast, our method utilizes the structured intermediate distributions of AIS, inspired by Sequential Monte Carlo (SMC), to achieve a more stable and accurate variational approximation. This approach effectively avoids weight collapse, particularly in high-dimensional scenarios.

\paragraph{Differentiable AIS}

Our method builds upon a well-established line of research~\cite{neal2001annealed, del2006sequential, zhang2021differentiable, xu2023embracing, chen2025dequantified}, and is specifically designed to overcome the known limitations of IW methods through AIS. We highlight the key differences between our approach and the Differentiable AIS (DAIS) method proposed by \cite{zhang2021differentiable}. DAIS circumvents the non-differentiability of traditional AIS by removing the Metropolis-Hastings correction, thereby enabling gradient-based optimization of the marginal likelihood. It has also been extended to black-box variational inference settings~\cite{jankowiak2022surrogate}. However, unlike our method, DAIS relies on a perturbed Hamiltonian system, whereas we adopt an inhomogeneous Unadjusted Langevin Algorithm (ULA). These represent fundamentally different formalisms: Hamiltonian mechanics typically employ symplectic integrators such as leapfrog methods, while Langevin dynamics utilize reverse stochastic differential equations (SDEs). Moreover, our algorithm is grounded in nonequilibrium statistical mechanics~\cite{nilmeier2011nonequilibrium} and is applied to Bayesian inference for Gaussian Process Latent Variable Models (GPLVM), in contrast to prior methods, which primarily focus on Bayesian linear regression.

\paragraph{Diffusion models} Our approach shares a fundamental connection with diffusion models \cite{ho2020denoising, song2020score, li2023scire} through the use of nonequilibrium statistical mechanics. While diffusion models, especially in generative modeling \cite{ruthotto2021introduction, croitoru2023diffusion}, use reverse stochastic processes to transform latent variables back to data space, they primarily focus on data generation rather than variational inference. In contrast to previous approaches, and similarly to \cite{xu2024sparse, xu2025bayesian}, our framework leverages the Unadjusted Langevin Algorithm (ULA) to better approximate posterior distributions by directly optimizing variational objectives.
 Additionally, our method models latent variable dynamics through forces driving the system toward equilibrium, drawing inspiration from nonequilibrium thermodynamics where systems relax to steady states via perturbative dynamics.
\begin{figure*}[t]
\centering
\includegraphics[width=\textwidth]{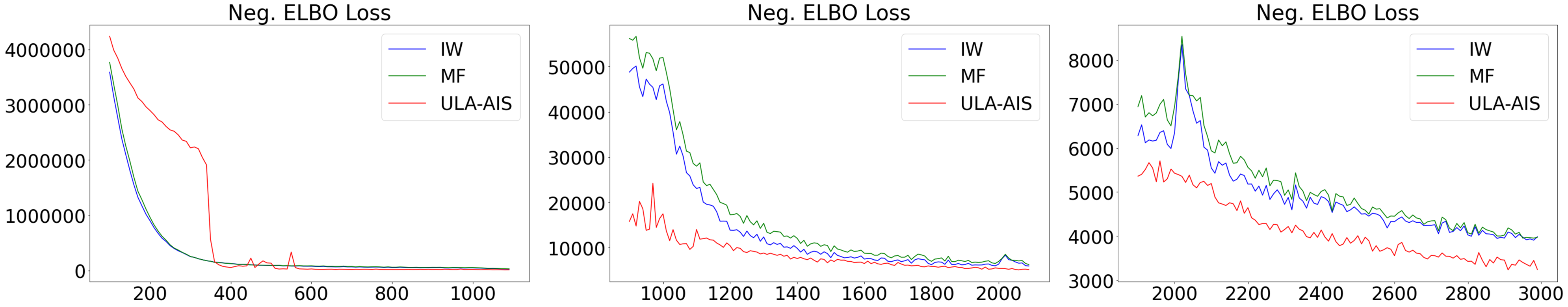}
\caption{The negative ELBO convergence curves of the three methods on the Frey Faces dataset. It is noted that as the number of iterations increase, the y-axis scale gradually increases from left to right.}
\label{fig:figure7}
\end{figure*}

\begin{table*}[ht]
  \centering
  \caption{Comparison of MF-GPLVM, IWVI-GPLVM, and VAIS-GPLVM under different number of iterations for two image datasets}
  \label{tab:comparison2}
  \resizebox{0.9\textwidth}{!}{
  \begin{tabular}{ccccccc}
    \toprule
    Dataset & Data Dim & Method & Iterations & Negative ELBO & MSE & Negative Expected Log Likelihood \\
    \midrule
    \multirow{9}{*}{Frey Faces} & \multirow{9}{*}{(1965,560)} & \multirow{3}{*}{MF-GPLVM} & 1000 & 48274 (443) & 468 (9) & 46027 (356) \\
       &    &  & 2000 & 6346 (20) & 95 (1) & 4771 (17) \\
       &    &  & 3000 & 3782 (15) & 69 (0.2) & 2822 (3) \\
       \cmidrule(lr){3-7}
       &  & \multirow{3}{*}{IWVI-GPLVM} & 1000 & 42396 (426) & 394 (8) & 39936 (312) \\
       &    & & 2000 & 5643 (15) & 76 (1) & 4292 (13) \\
       &    &  & 3000 & 3596 (14) & 63 (0.5) & 2535 (4) \\
       \cmidrule(lr){3-7}
       &  & \multirow{3}{*}{VAIS-GPLVM (ours)} & 1000 & \textbf{12444} (451) & \textbf{121} (9) & \textbf{10543} (322) \\
       &    &  & 2000 & \textbf{5031} (16) & \textbf{66} (1) & \textbf{3130} (15) \\
       &    &  & 3000 & \textbf{3249} (12) & \textbf{57}(0.3)  & \textbf{2226} (3) \\
       \midrule
    \multirow{3}{*}{MNIST} & \multirow{3}{*}{(2163,784)} & MF-GPLVM & 2000 & -432.32(0.33) & 0.27(0.004) & -552.87(0.28) \\
       
       \cmidrule(lr){3-7}
       & & IWVi-GPLVM & 2000 & -443.64(0.37) & \textbf{0.25}(0.003) & -567.13(0.31) \\
       
       \cmidrule(lr){3-7}
       &  & VAIS-GPLVM (ours)& 2000 & \textbf{-453.18}(0.27) &  \textbf{0.25}(0.002)& \textbf{-569.93}(0.26) \\
      
    \bottomrule
  \end{tabular}}
\end{table*}

\begin{figure}[t]
\centering
\includegraphics[width=\linewidth]{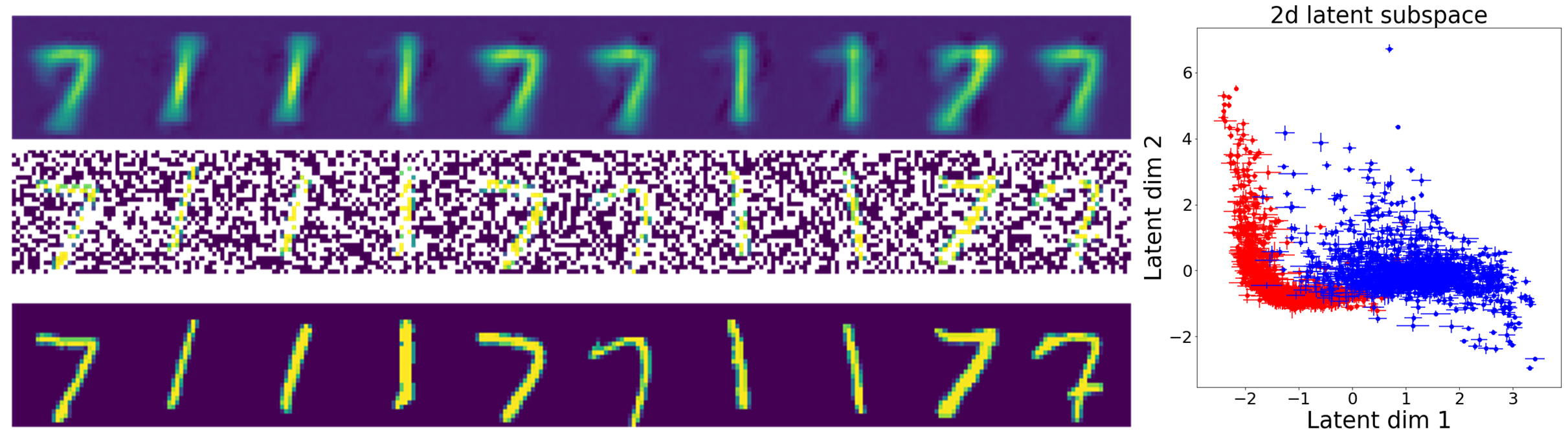}
\caption{For MNIST with 75\% missing pixels, we used digits 1 and 7. The bottom row shows ground truth, while the top row shows reconstructions from the 5D latent space. The 2D plot on the right visualizes the dimensions with the smallest lengthscales.}
\label{fig:figure6}
\end{figure}

\section{Experiments}
\subsection{Baseline Methods}
In the following section, we present two sets of experiments. In the first set of experiments, our aim is to demonstrate the quality of our model in unsupervised learning tasks such as data dimensionality reduction and clustering. This will allow us to evaluate the ability of our model to preserve the original information in the data. In the second set of experiments, we evaluate the expressiveness and efficiency of our model on the task of image data recovery. 

We compare three different approaches: (a) Classical Sparse VI based on mean-field (MF) approximation \cite{titsias2010bayesian}; (b) Importance-weighted (IW) VI \cite{salimbeni2019deep}; (c) The Unadjusted Langevin Diffusion Variational AIS model (hereinafter referred to as VAIS-GPLVM) is defined by the algorithm proposed in this paper. We also provide guidelines on how to tune the step sizes and annealing schedules in Algorithm \ref{algorithm} to optimize performance. We conducted all our experiments on a Tesla A100 GPU.

\subsection{Dimensionality
Reduction}

The multi-phase Oilflow data \cite{bishop1993analysis} consists of 1000, 12d data points
belonging to three classes which correspond to the
different phases of oil flow in a pipeline. We reduced the data dimensions to 10 while attempting to preserve as much information as possible. We  report the reconstruction error and MSE with  ±2
 standard errors over ten optimization runs. Since
the training is unsupervised, the inherent ground-truth
labels were not a part of training. The 2d projections of the latent space for oilflow data clearly shows that our model is able to discover the class structure.  

To highlight the strength of our model, we set the same  experimental hyperparameters and compare the learning curves of two state-of-the-art models. The results are shown in Fig. \ref{fig1}. We also tested our model performance on another toy dataset, Wine Quality \cite{cortez2009modeling}, where we used the white variant of the Portuguese "Vinho Verde" wine. From table \ref{tab:comparison}, we observe that after sufficient training, our proposed method yields lower reconstruction loss and MSE than IWVI and MF methods. It is noted that our proposed method does not show an increase in time complexity compared to the baseline method IW. Therefore, even though we used a fixed number of iterations, we can ensure the fairness of the experiments.

\subsection{Make Predictions in Unseen Data}

We conducted reconstruction experiments on MNIST and Frey Faces to assess model uncertainty under missing structured inputs. For MNIST, we used digits 1 and 7 with a 5-dimensional latent space; each image is 784- dimensional. For Frey Faces \cite{roweis2000nonlinear}, we used the full dataset of 1965 images ($20\times28$ pixels, 560-dimensional) with a 20-dimensional latent space. In both datasets, 5\% of training samples had 75\% of their pixels removed to test reconstruction. Results, shown in Fig. \ref{fig:figure5} and Fig. \ref{fig:figure6}, reflect sampling from the learned latent distributions. Our setup follows prior work by \cite{titsias2010bayesian} and \cite{gal2014distributed}. Additional details are provided in the Appendix.

To demonstrate the effectiveness of our method in producing more accurate likelihoods and tighter variational bounds on image datasets, we present in Table \ref{tab:comparison2} the negative ELBO, negative log-likelihood, and mean squared error (MSE) for reconstructed images on the Frey Faces and MNIST datasets, comparing with state-of-the-art methods. Our results show that our method achieves lower variational bounds and converges to higher likelihoods, indicating superior performance in high-dimensional and multi-modal image data. This suggests that adding Langevin transitions appears to improve the convergence of the traditional VI methods.

We also present in Fig. \ref{fig:figure7} a comparison of the negative ELBO convergence curves for Frey Faces datasets between our method and two other state-of-the-art methods. To better illustrate our lower convergence values, we gradually increase the y-axis scale from left to right. An interesting observation is that, compared to the IW and MF methods, our proposed method sometimes exhibits sudden drops in the loss curve, as shown in the leftmost plot of Fig. \ref{fig:figure7}. This can be attributed to the fact that, by adding Langevin transitions, the algorithm's variational distribution gradually moves from the current distribution towards the true posterior distribution, resulting in sudden drops in the loss function when reaching the target distribution. Thus, such phenomena can be regarded as a common feature of annealed importance sampling and it becomes even more obvious in high-dimensional datasets.

\subsection{Effective Sample Size (ESS) Analysis}

In our experiments, we observed clear evidence of weight collapse in IW-GPLVM, particularly as the dimensionality of the latent space increases. Below, we present additional results from the Brendan Faces reconstruction task, comparing IW-GPLVM and our proposed VAIS-GPLVM using standard diagnostic metrics:

\begin{table}
\centering
\caption{Comparison of ESS and Weight Entropy for IWVI-GPLVM and VAIS-GPLVM (Ours) on the Brendan Faces Reconstruction Task. VAIS-GPLVM demonstrates a significant improvement in both Effective Sample Size (ESS) and Weight Entropy, indicating that it mitigates sample collapse and promotes a more uniform weight distribution.}
\label{ess}
\resizebox{0.48\textwidth}{!}{\begin{tabular}{|c|c|c|}
\hline
\textbf{Metric} & \textbf{IWVI-GPLVM} & \textbf{VAIS-GPLVM (Ours)}  \\
\hline
ESS ($K=25$) & 4.1 & 20.3 \\
\hline
Weight Entropy ($K=25$) & 0.9 & 2.6  \\
\hline
\end{tabular}}
\end{table}

Effective Sample Size (ESS) \cite{rainforth2018tighter} quantifies the number of samples that effectively contribute to the final estimate, despite using all $K$ particles. It is defined as

\begin{align}
\text{ESS} = \frac{1}{\sum_{k=1}^{K} \tilde{w}_k^2},
\end{align}

where $\tilde{w}_k$ are the normalized importance weights. A low ESS indicates that only a few particles dominate the estimate, reflecting weight collapse.

Weight Entropy is defined as

\begin{align}
H(\tilde{w}) = -\sum_{k=1}^{K} \tilde{w}_k \log \tilde{w}_k,
\end{align}

which measures the dispersion of the importance weights. Higher entropy suggests a more uniform distribution of weights and better utilization of available samples.

As shown in the Table \ref{ess}, VAIS-GPLVM achieves substantially higher ESS and weight entropy, indicating more diverse and stable sampling behavior. In contrast, IW-GPLVM suffers from severe weight concentration, corroborating its known theoretical limitations and aligning with our earlier motivation that IWVI tends to experience weight collapse in high-dimensional settings.

\section{Conclusion}
In this paper, we propose VAIS-GPLVM, a novel variational approach for GPLVMs based on Annealed Importance Sampling. By leveraging annealing and unadjusted Langevin dynamics, our method estimates the ELBO via a sequence of tractable intermediate distributions. Empirical results on high-dimensional and structured datasets demonstrate improved variational bounds, faster convergence, and greater robustness. Notably, sharp drops in the loss curve further validate the effectiveness of our approach. Overall, VAIS-GPLVM offers a promising direction for variational learning in latent variable models.

\newpage
\section*{Acknowledgement}
This work was supported by the Fundamental Research Program of Guangdong, China (Grant No. 2023A1515011281). We would also like to express our sincere gratitude to the three reviewers and the meta-reviewers for their thorough and constructive feedback, which significantly helped improve the quality of this paper.

\bibliography{uai2025-template}

\newpage

\onecolumn

\appendix

\section{Derivation of Equation (3) and (5)}
\subsection{Derivation of Equation (3)}

First, decompose the log evidence into a double summation form along the observation dimensions: \begin{equation} \log p(X) = \sum_{n=1}^{N} \sum_{d=1}^{D} \log p(x_{n,d}) \end{equation}

Term-wise Application of Jensen's Inequality
Apply Jensen's inequality to each term $\log p(x_{n,d})$ and introduce the variational distribution to obtain: \begin{equation}\log p(x_{n,d}) \geq \mathbb{E}_{q(f_d,u_d)q(h_n)}[\log p(x_{n,d}|f_d,h_n)] - \text{KL}(q(h_n)|p(h_n)) - \text{KL}(q(f_d, u_d)|p(f_d, u_d)) \end{equation} 
where $q(f_d,u_d)=p(f_d|u_d)q(u_d)$.

Sum the bounds of all terms to obtain the initial variational lower bound: \begin{equation}\label{i} \text{MF-ELBO}_f(\gamma,\psi) = \sum_{n,d} \left[ \mathbb{E}_{q(f_d,u_d)q(h_n)}[\log p(x_{n,d}|f_d,h_n)] - \text{KL}(q(h_n)|p(h_n)) - \text{KL}(q(f_d, u_d)|p(f_d, u_d)) \right]  \end{equation}

Sparse Variational Approximation: \begin{equation} q(f_d) = \int p(f_d|u_d)q(u_d)du_d \end{equation}
The KL term can then be simplified as: \begin{equation} \text{KL}(q(f_d, u_d)|p(f_d, u_d)) = \int p(f_d|u_d) q(u_d)\log \frac{ p(f_d|u_d) q(u_d)}{p(f_d|u_d)p(u_d)}du_d df_d=\text{KL}(q(u_d)|p(u_d)) \end{equation} and we have, \begin{equation}
\label{i1}
\mathbb{E}_{q(f_d,u_d)q(h_n)}[\log p(x_{n,d}|f_d,h_n)]=\mathbb{E}_{q(f_d)q(h_n)}[\log p(x_{n,d}|f_d,h_n)] \end{equation}

Substitute the simplified KL term and Equation (\ref{i1}) into Equation (\ref{i}) to obtain the MF-ELBO consistent with the main  text:  \begin{equation} \text{MF-ELBO}(\gamma,\psi) = \sum_{n,d} \left[ \mathbb{E}_{q(f_d)q(h_n)}[\log p(x_{n,d}|f_d,h_n)] - \text{KL}(q(h_n)|p(h_n)) - \text{KL}(q(u_d)|p(u_d)) \right]  \end{equation}

\subsection{Derivation of Equation (5)}
Following the IWAE approach, we apply importance-weighted variational inference to the latent variable $h_n$ (the initial steps align with mean-field variational inference and are thus omitted). The MF-ELBO $\text{MF-ELBO}(\gamma,\psi)$ is rewritten in the previous step as: \begin{equation} \sum_{n,d} \left[ \mathbb{E}_{q} \left[ \log \left( \frac{p(x_{n,d} \mid f_d, h_{n}) p(h_{n})}{q(h_{n})} \right) \right] - \text{KL}(q(u_d) | p(u_d)) \right]. \end{equation}

Sampling: Independently draw $K$ samples $h_{n,1}, \dots, h_{n,K}$ from $q(h_n)$.
Estimation Construction: Approximate the likelihood term using importance weighting: \begin{equation} \frac{p(x_{n,d} \mid f_d, h_{n}) p(h_{n})}{q(h_{n})} \approx \frac{1}{K} \sum_{k=1}^K \frac{p(x_{n,d} \mid f_d, h_{n,k}) p(h_{n,k})}{q(h_{n,k})} . \end{equation} This estimator satisfies consistency in expectation: \begin{equation}  \mathbb{E}_{q(h_{n,1:K})} \left[ \frac{1}{K} \sum_{k=1}^K \frac{p(x_{n,d} \mid f_d, h_{n,k}) p(h_{n,k})}{q(h_{n,k})} \right] = p(x_{n,d} \mid f_d). \end{equation}  When $K=1$, it reduces to the mean-field case:  \begin{equation} \mathbb{E}_{q(h_{n,1})} \left[ \frac{p(x_{n,d} \mid f_d, h_{n,1}) p(h_{n,1})}{q(h_{n,1})} \right] = p(x_{n,d} \mid f_d).  \end{equation}

Substitute the importance-weighted estimator into the ELBO  to obtain Equation 5: \begin{equation} \log p(x_{n,d}) \geq E_{q(f_d)q(h_n)} \left[ \log \left( \frac{1}{K} \sum_{k=1}^K \frac{p(x_{n,d} \mid f_d, h_{n,k}) p(h_{n,k})}{q(h_{n,k})} \right) \right] - \text{KL}(q(u_d) | p(u_d)).  \end{equation} This is the IW-ELBO in Equation (5) with $\sum_{n,d}$ .

\section{Derivation of the Overdamped Langevin Path Probability Ratio}

For ease of sampling, we define the corresponding Euler-Maruyama discretization as,
\begin{align}
\label{ula1}
    \mathbf{H}_{k}=\mathbf{H}_{k-1}+\eta \nabla \log q_{k}\left(\mathbf{H}_{k-1}\right)+\sqrt{2 \eta} \boldsymbol{\epsilon} _{k-1},
\end{align}
where $\boldsymbol{\epsilon}_k \sim \mathcal{N}(0,I)$.
Based on results by \cite{nilmeier2011nonequilibrium}, the backward step is realized by
\begin{align}
\mathbf{H}_{k-1}=\mathbf{H}_k+\eta \nabla \log q_k\left( \mathbf{H}_k \right) +\sqrt{2\eta}\tilde{\boldsymbol{\epsilon}}_{k-1},
\end{align}
Thus we have,
\begin{align}
\begin{aligned}
  \eta \nabla \log q_k\left( \mathbf{H}_{k-1} \right) +\sqrt{2\eta}\boldsymbol{\epsilon }_{k-1}&=-\eta \nabla \log q_k\left( \mathbf{H}_k \right) -\sqrt{2\eta}\boldsymbol{\tilde{\epsilon}}_{k-1}   
\end{aligned}
\end{align}
Then,
\begin{align}
\boldsymbol{\tilde{\epsilon}}_{k-1}=-\sqrt{\frac{\eta}{2}}\left( \nabla \log q_k\left( \mathbf{H}_{k-1} \right) +\nabla \log q_k\left( \mathbf{H}_k \right) \right) -\boldsymbol{\epsilon }_{k-1}
\end{align}
Finally,
\begin{align}
\begin{aligned}
 \log \frac{\mathcal{T} _k\left( \mathbf{H}_k\mid \mathbf{H}_{k-1} \right)}{\tilde{\mathcal{T}}_k\left( \mathbf{H}_{k-1}\mid \mathbf{H}_k \right)}&=\log \frac{p\left( \boldsymbol{\epsilon }_{k-1} \right) \left| \det \left( \frac{\partial \mathbf{H}_k}{\partial \boldsymbol{\epsilon }_{k-1}} \right) \right|}{p\left( \boldsymbol{\tilde{\epsilon}}_{k-1} \right) \left| \det \left( \frac{\partial \mathbf{H}_{k-1}}{\partial \boldsymbol{\tilde{\epsilon}}_{k-1}} \right) \right|}\\&=\log \frac{p\left( \boldsymbol{\epsilon }_{k-1} \right)}{p\left( \boldsymbol{\tilde{\epsilon}}_{k-1} \right)}\\&=\frac{1}{2}\left( \left\| \boldsymbol{\tilde{\epsilon}}_{k-1} \right\| ^2-\left\| \boldsymbol{\epsilon }_{k-1} \right\| ^2 \right) 
\end{aligned}
\end{align}

\section{A Stochastic Variant of VAIS-GPLVM}

Instead of computing the gradient of the full log likelihood, we suggest to use a stochastic variant to subsampling datasets into a mini-batch $\mathcal{D}_J$ with $\left | \mathbf{X}_J\ \right | =B$, where $J\subset \{1,2,..,N\}$ is the indice of
any mini-batch. We can thus define an estimator of  $\nabla \log p(\mathbf{X} \mid \cdot)$ in Eq. (12) as,
\begin{align}
    \nabla \log p(\mathbf{X} \mid \cdot)\approx \frac{N}{B} \nabla\log  p(\mathbf{X}_J \mid \cdot)
\end{align}
In the meantime, we replace the $p\left(\mathbf{X}, \mathbf{H}_{K}\right)$ term in Eq. (7) with another estimator computed using an independent mini-batch of indices  $I\subset \{1,2,..,N\}$
with $\left | \mathbf{X}_I\ \right | =B$, $i.e.$
\begin{align}
    p\left(\mathbf{X}, \mathbf{H}_{K}\right)\approx p\left( \mathbf{H}_{K}\right)p\left(\mathbf{X}_I\mid \mathbf{H}_{K}\right)^\frac{N}{B} 
\end{align}
With  jointly using the reparameterization trick and stochastic gradient descent, we finally derive a  stochastic variant of the Stochastic Unadjusted Langevin Diffusion AIS algorithm for the LVGP models as describe in Algorithm
1. Thanks to GPU acceleration, we can extend the proposed algorithm to larger datasets, such as image-based visual tasks.

\section{Practical Guidelines}
In the context of this paper, the posterior distribution refers to the distribution of the latent variables given the observed data. This distribution is often intractable and challenging to sample from directly. VAIS-GPLVM aims to approximate this posterior distribution by transforming it into a sequence of intermediate distributions, which can be more tractable and easier to sample from.

The annealing process gradually transforms the posterior distribution by introducing a temperature parameter $\beta$ . By annealing from $\beta=0$  to $\beta=1 $ , we move from an initial distribution, where the posterior is approximated by a simpler distribution to the target posterior distribution itself. The key idea behind annealing is it allows for a smoother exploration of the posterior space. At each intermediate distribution, we can use importance sampling to estimate the evidence by sampling from the proposal distribution and reweighting the samples using the ratios of the target and proposal distributions.

As the annealing process progresses, the samples from the proposal distribution gradually become more representative of the target distribution. This means that the exploration of the posterior space is not limited to a specific region but covers a wider range of possible configurations of the latent variables.

The benefit of this exploration is that it allows for a more accurate estimation of the evidence, which corresponds to a tighter lower bound in the variational learning framework. By gradually annealing the temperature and exploring different distributions, VAIS-GPLVM can capture more complex structures in the posterior distribution, leading to better variational approximations in complex data and high-dimensional spaces.

When using the Unadjusted Langevin Diffusion method for sampling, one key challenge is to determine an appropriate step size $\eta_k$ A fixed step size may work well for some samples but may be suboptimal for others. To address this issue, we can use the Adagrad \cite{kingma2014adam} optimizer to adaptively adjust the step size based on the historical gradient information. Specifically, for each dimension of the sampled variables, we divide the initial step size by the square root of the sum of squared gradient values for that dimension up to a noise. This technique can help achieve better performance and faster convergence, especially when dealing with complex and high-dimensional distributions where finding an appropriate step size is challenging. The adaptive step size adjustment can be implemented in combination with other techniques, such as early stopping, to further improve the sampling efficiency.

$$\eta_k = 0.9\eta_{k-1}+0.1\dfrac{\eta_0}{\sqrt{G_{k} +\epsilon}}$$

where $G_k$ is the sum of squared gradient values up to step $k$ in Eq. (17), $\epsilon$ is a small smoothing term to avoid division by zero,and $\eta_0$ is the initial step size.

In the context of Annealed Importance Sampling (AIS), choosing an optimal temperature schedule ${\beta_k}$ is a challenging task. When choosing an appropriate annealing schedule for Stochastic Gradient Annealed Importance Sampling, there are several trade-offs and considerations to keep in mind:
\begin{itemize}
    \item Computational Efficiency: The annealing schedule should be carefully designed to balance the computational resources required for estimating the evidence. Too many bridging densities can lead to excessive computational burden, while too few densities may result in less accurate estimates.
    \item Exploration vs Exploitation: The annealing schedule should strike a balance between exploration and exploitation of the posterior distribution. An aggressive schedule that moves quickly from the base distribution to the posterior may lead to exploration limitations, while a slow schedule may lead to insufficient exploration and inefficiency.
    \item Smoothness of Transition: The annealing schedule should ensure a smooth transition between bridging densities. Abrupt changes in the densities can result in high-variance importance weights, which may lead to inaccurate estimates. Smooth transitions can be achieved by gradually adjusting the temperature or using appropriate interpolation functions.
\end{itemize}
We often use a linear schedule, where the temperature values are fixed and regularly spaced between 0 and 1. However, this approach may not always work well in practice, as the search space is complex and high-dimensional. 

Alternatively, we can try to  learn the temperature values ${\beta_k}$ directly as additional inference parameters $\phi$. This can be done using various techniques, such as gradient-based optimization. By doing so, we can obtain a temperature schedule that is tailored to the specific problem at hand and achieve better sampling performance. Additional experimental information can be seen in Table \ref{tab:five-by-eight}.

\begin{table*}[h]
\centering
\resizebox{0.9\textwidth}{!}{\begin{tabular}{|c|c|c|c|c|c|c|c|}
\hline
Dataset & Task & $N$ & $D$ & $Z$ & $Q$ & LR & $K$  \\
\hline
Oilflow & Dimensionality Reduction & 1000 & 12 & 50 & 10 & 0.02 & 5 \\
\hline
Wine Quailty & Dimensionality Reduction & 1599 & 11 & 50 & 9 & 0.02 & 5 \\
\hline
Frey Face  & Missing Data Recovery & 1965 & 560 & 50 & 20 & 0.02 & 25 \\
\hline
MNIST & Missing Data Recovery & 2163 & 784 & 50 & 5 & 0.02 & 25 \\
\hline
\end{tabular}}
\caption{Training experimental configuration where $N$ and $D$ denote the number of data points and data space
dimensions, $Z$ denotes the number of inducing inputs shared across dimensions, $Q$ denotes the dimesionality of
the latent space, LR denotes the learning rate, $K$ denotes the length  of the transition chain in VAIS-GPLVM and in IW $K$ denotes the number of repetitions of sampling .}
\label{tab:five-by-eight}
\end{table*}

\section{Details for Implementing on Missing Data Tasks}
Specially, our training procedure leverages the
marginalisation principle of Gaussian distributions
and the fact that the data dependent terms of
the  ELBO factorise across data points and
dimensions. This means we can trivially marginalise
out the missing dimensions $\boldsymbol{x}_a$, because each individual data point $\boldsymbol{x}$ is modelled as a joint Gaussian.
Consider a high-dimensional point $\boldsymbol{x}$ which we split
into observed, $\boldsymbol{x}_o$ and unobserved $\boldsymbol{x}_a$ dimensions,
\begin{equation}
    \int \prod_{d \in a} \prod_{d \in o} p\left(\boldsymbol{x}_{a}, \boldsymbol{x}_{o} \mid \boldsymbol{f}_{d}, \mathbf{H}\right) d \boldsymbol{x}_{a}=\prod_{d \in o} p\left(\boldsymbol{x}_{o} \mid \boldsymbol{f}_{d}, \mathbf{H}\right)
\end{equation}

\begin{figure}[ht]
\centering
\begin{minipage}{0.48\linewidth}
    \centering
    \includegraphics[width=\linewidth]{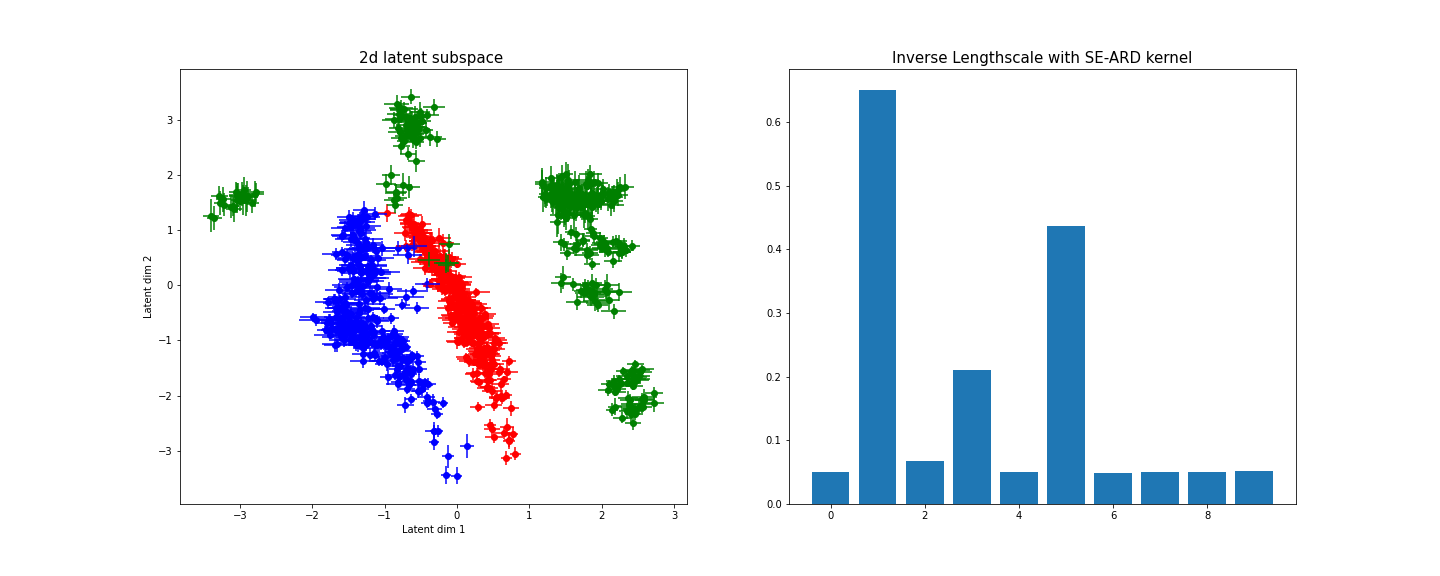}
    \caption{Dimensionality  Reduction Results for MF method.}
\end{minipage}
\hfill
\begin{minipage}{0.48\linewidth}
    \centering
    \includegraphics[width=\linewidth]{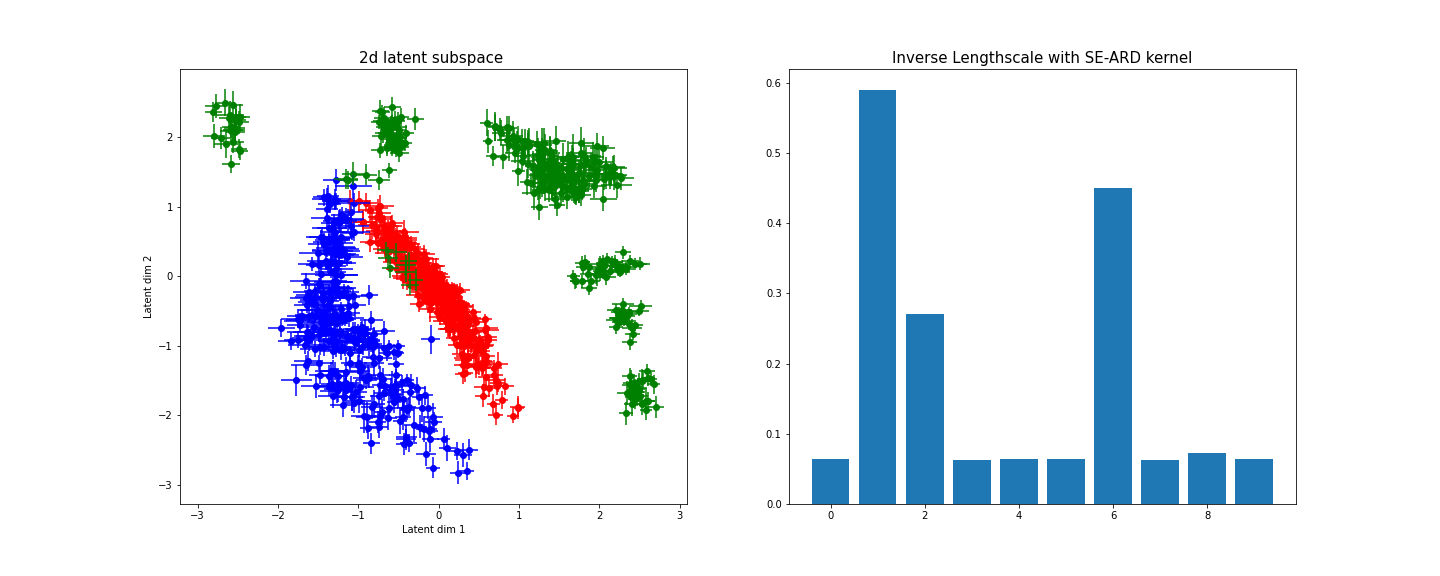}
    \caption{Dimensionality Reduction Results for IW method.}
\end{minipage}
\end{figure}

\begin{figure*}[ht]
\centering
\begin{minipage}{0.48\linewidth}
    \centering
    \includegraphics[width=\linewidth]{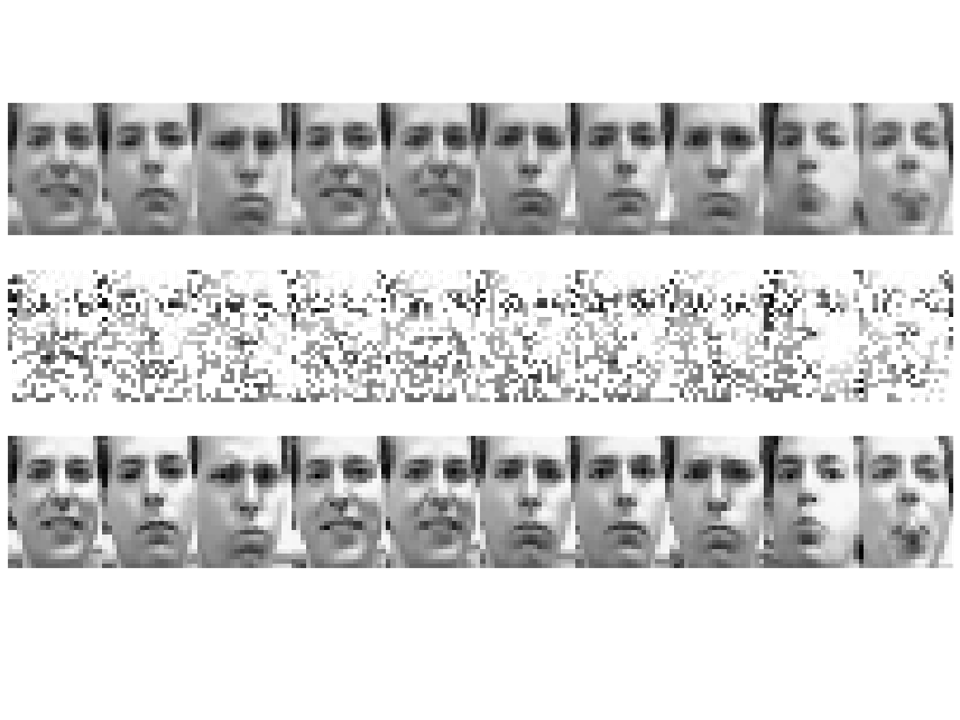}
    \caption{Missing Data Recovery Results for MF method. The bottom row represents the ground truth data and the top row showcases the reconstructions from the 20-dimensional latent distribution.}
\end{minipage}
\hfill
\begin{minipage}{0.48\linewidth}
    \centering
    \includegraphics[width=\linewidth]{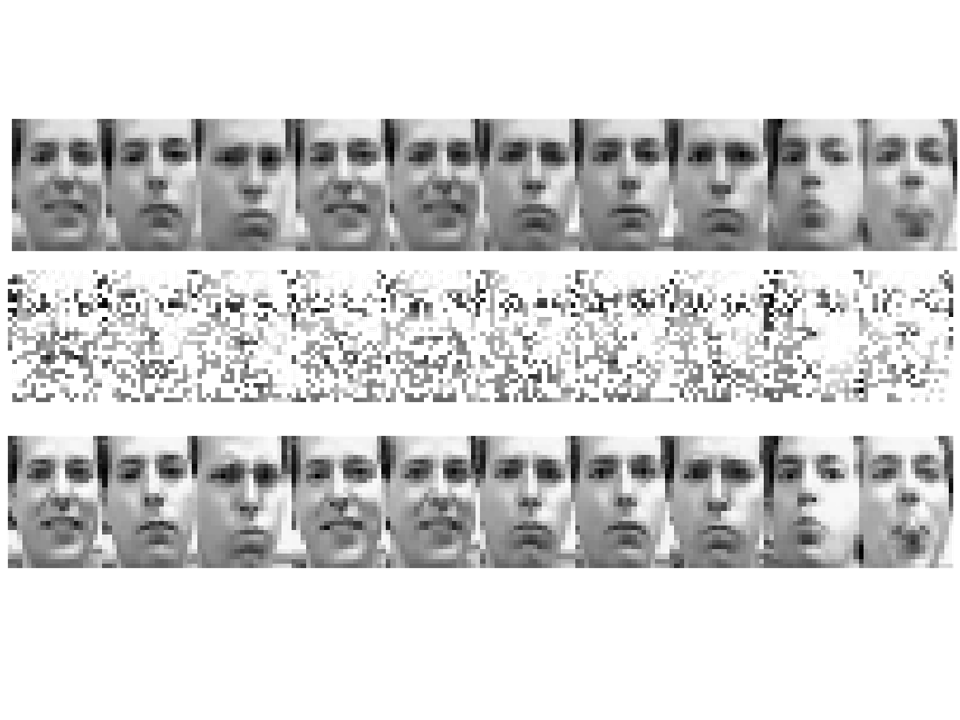}
    \caption{Missing Data Recovery Results for IW method. The bottom row represents the ground truth data and the top row showcases the reconstructions from the 20-dimensional latent distribution.}
\end{minipage}
\end{figure*}

In this formula, the indices of missing and observed dimensions are denoted by $a$ and $o$ respectively, where $D = a \cup o$ represents all dimensions in the data. The marginal distributions $\boldsymbol{f}_{d} \in \mathbb{R}^N$ are defined in Eq. (4).The latent variables $\boldsymbol{h}_n$ for each data point are informed only by the observed dimensions. Furthermore, we can easily reconstruct the missing dimensions during training by constructing a variational latent distribution $q(\mathbf{H})$, as described in Section 4. This approach enables us to efficiently handle missing dimensions in high-dimensional datasets without requiring major modifications to the overall training process.

\subsection{Compared to Standard GPLVM}
We have also conducted additional experiments comparing our proposed approach to the Standard GPLVM \cite{lawrence2003gaussian} in Table \ref{tab:st}. We performed experiments 10 times and averaged the results, analyzing the performance (in terms of MSE and NLL) on four different datasets. Due to limited computational resources, we were only able to run the Standard GPLVM on a subset of the image datasets. For the image reconstruction task, we randomly selected 300 images as the training set and used consistent hyperparameters for the other experiments.
\begin{table*}[ht]
\centering

\resizebox{0.9\textwidth}{!}{
\begin{tabular}{|c|c|c|c|}
\hline
\textbf{Dataset (Size)} & \textbf{Method} & \textbf{MSE} & \textbf{NLL} \\
\hline
\multirow{2}{*}{Oilflow (1000, 12)} 
& Standard GPLVM & 2.45 (0.05) & -12.42 (0.07) \\
& VAIS-GPLVM (Ours)      & \textbf{1.71} (0.04) & \textbf{-15.81} (0.04) \\
\hline
\multirow{2}{*}{Wine Quality (1599, 11)} 
& Standard GPLVM & 30.53 (0.03) & 2.82 (0.02) \\
& VAIS-GPLVM (Ours)      & 30.79 (0.04) & \textbf{2.42} (0.03) \\
\hline
\multirow{2}{*}{Frey Faces (300, 560)} 
& Standard GPLVM & 130.00 (7.00) & 2632.00 (6.00) \\
& VAIS-GPLVM (Ours)      & \textbf{115.00} (6.00) & \textbf{2417.00} (5.00) \\
\hline
\multirow{2}{*}{MNIST (300, 784)} 
& Standard GPLVM & 0.36 (0.01) & -484.00 (3.00) \\
& VAIS-GPLVM (Ours)      & \textbf{0.31} (0.01) & \textbf{-496.00} (2.00) \\
\hline
\end{tabular}
}
\caption{Comparison of MSE and NLL between Standard GPLVM and our VAIS-GPLVM across four datasets.}
\label{tab:st}
\end{table*}

\subsection{Runtime Analysis}

We observed that the runtime of Importance-Weighted (IW) VI and VAIS-GPLVM increases almost linearly with $K$. For IW, this is due to the $K$ repeated samplings of latent variables, each with a complexity of $O(nm^2)$ from the GPLVM model. As $K$ increases, the repeated samplings dominate the runtime. In contrast, VAIS-GPLVM requires only one such sampling, with additional computations focused on the lighter Langevin stochastic flow during annealing. As shown in Table \ref{tab:runtime_comparison}, AIS becomes more efficient than IW when $K$ exceeds a certain threshold on the Frey Faces dataset.

\begin{table}[ht]
\centering
\resizebox{0.9\textwidth}{!}{\begin{tabular}{|c|c|c|c|c|c|}
\hline
\textbf{Method} & $K=5$ & $K=10$ & $K=15$ & $K=20$ & $K=25$ \\
\hline
IWVI-GPLVM         & 1.46s & 2.85s & 4.06s & 5.45s & 7.03s \\
\hline
VAIS-GPLVM (Ours) & 1.53s & 2.65s & 3.79s & 4.80s & 5.93s \\
\hline
\end{tabular}}
\caption{Comparison of running time between IWVI-GPLVM and VAIS-GPLVM  in one epoch for Frey Faces}
\label{tab:runtime_comparison}
\end{table}

\subsection{Additional Results}
In this section, we will demonstrate the visual effects of the MF and IW methods on three datasets: Oilflow, MINIST, and Frey Faces. These visualizations will be used for comparison with the main text. There results can be seen in Fig. 6, Fig.7, Fig.8, Fig.9, Fig.10, Fig.11.

\begin{figure*}[ht]
\centering
\begin{minipage}{0.48\linewidth}
    \centering
    \includegraphics[width=\linewidth]{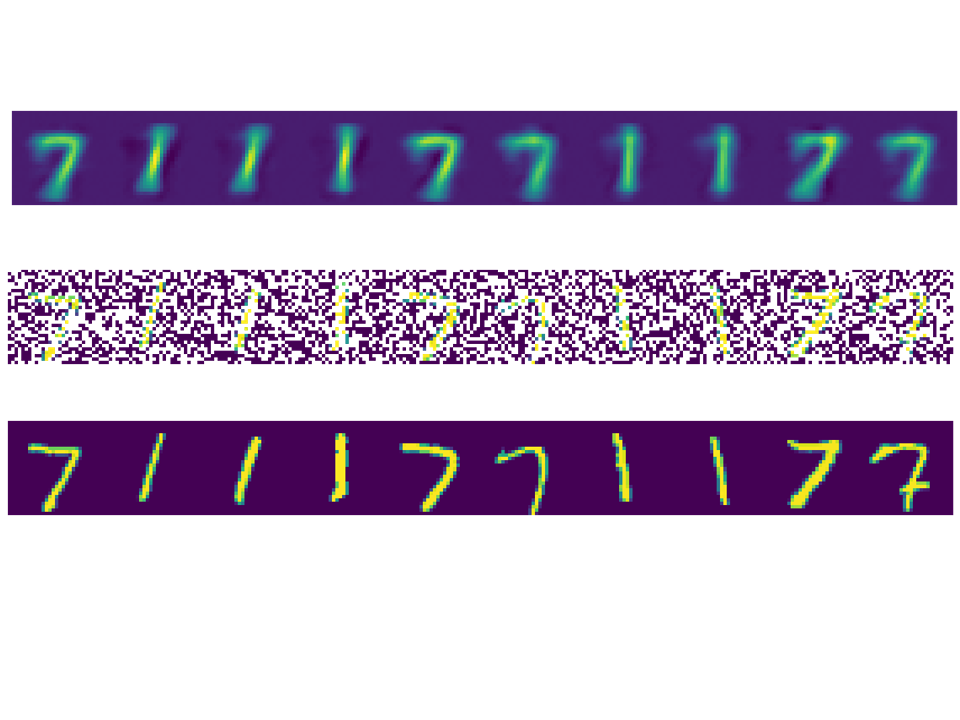}
    \caption{Missing Data Recovery Results for MF method. The top row represents the ground truth data and the bottom row showcases the reconstructions from the 5-dimensional latent distribution.}
\end{minipage}
\hfill
\begin{minipage}{0.48\linewidth}
    \centering
    \includegraphics[width=\linewidth]{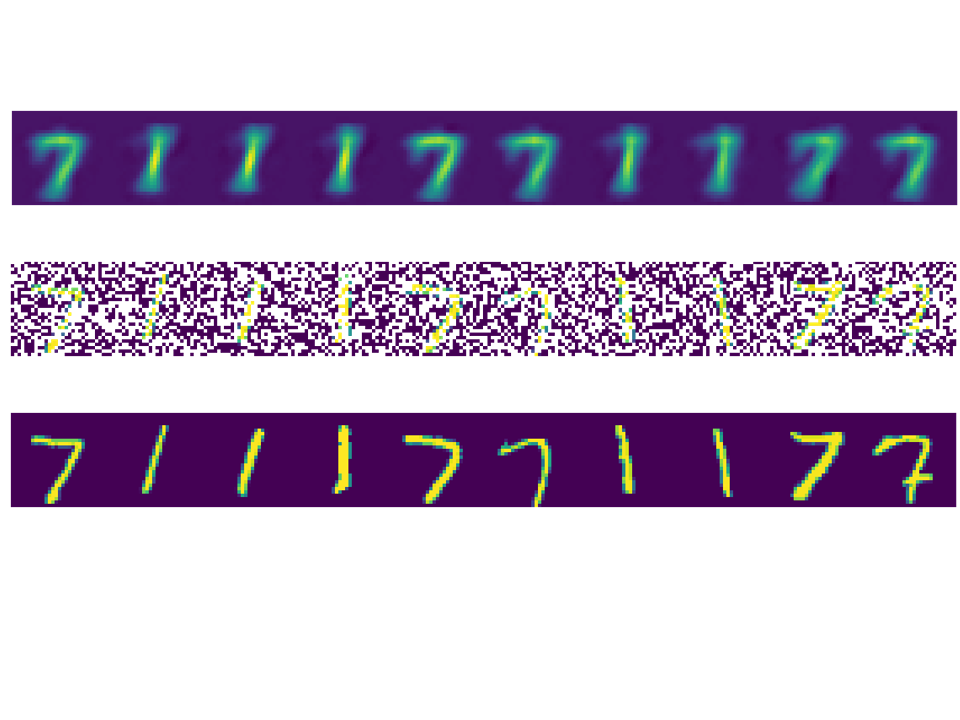}
    \caption{Missing Data Recovery Results for IW method. The top row represents the ground truth data and the bottom row showcases the reconstructions from the 5-dimensional latent distribution.}
\end{minipage}
\end{figure*}

From the visual appearance, it may seem that all three methods produce similar reconstructions. However, upon closer inspection, we can observe differences in certain details such as brightness and contrast. While these differences may be difficult to discern with the naked eye, we have quantified them using the mean squared error (MSE) between the reconstructed images and the ground truth. The MSE results for all three methods on the test set are reported in Tables 2 in the main text.

\section{Limitations and Future work}
One potential limitation could be the scalability of the method. As the size of the dataset increases, the computational resources required for estimating the evidence using VAIS-GPLVM may become more demanding. This is particularly true for large-scale datasets such as ImageNet, which contain millions of images. Running experiments on such massive datasets might pose challenges in terms of computational efficiency and memory requirements. Given that ImageNet involves higher-dimensional data, it may be more appropriate to combine GPLVM with other deep learning tools, such as convolutional neural networks (CNNs) \cite{li2021survey, he2020resnet} and transformers \cite{vaswani2017attention, lin2022survey}. Broader application scenarios are currently being explored to incorporate these tools effectively and leave room for future work.

Additionally, the annealing schedule plays a crucial role in the exploration of the posterior distribution. Designing an appropriate annealing schedule may require domain knowledge or trial and error experimentation. It might be necessary to tune the schedule to ensure a balance between exploration and exploitation, as well as a smooth transition between bridging densities.

Regarding the applicability of VAIS-GPLVM in real-world applications, its performance may depend on the specific characteristics and requirements of the domain. Different datasets and applications may exhibit unique challenges, such as data sparsity, high dimensionality, or non-linear relationships, which could affect the effectiveness of SG-AIS. Evaluating the performance of SG-AIS in different domains and addressing these challenges would require further experimentation and investigation.

\end{document}